\documentclass[10pt,twocolumn,letterpaper]{article}

\usepackage[pagenumbers]{cvpr} 

\usepackage{graphicx}
\usepackage{amsmath}
\usepackage{amssymb}
\usepackage{booktabs}
\usepackage{multirow}
\usepackage{url}

\usepackage[pagebackref,breaklinks,colorlinks]{hyperref}

\usepackage[capitalize]{cleveref}
\crefname{section}{Sec.}{Secs.}
\Crefname{section}{Section}{Sections}
\Crefname{table}{Table}{Tables}
\crefname{table}{Tab.}{Tabs.}

\def\confName{CVPR}
\def\confYear{2023}

\begin{document}

\title{StepFormer: Self-supervised Step Discovery and Localization \\ in Instructional Videos}

\author{Nikita Dvornik$\mathbf{^{1}}$
\quad
Isma Hadji$\mathbf{^1}$
\quad
Ran Zhang$\mathbf{^1}$
\quad
Konstantinos G. Derpanis$\mathbf{^{1,2}}$\\
\quad
Animesh Garg$\mathbf{^{3}}$
\quad
Richard P. Wildes$\mathbf{^{1,2}}$
\quad
Allan D. Jepson$\mathbf{^1}$\vspace{1pt}
\quad\\
{\large $\mathbf{^1}$Samsung AI Centre Toronto, $\mathbf{^2}$York University $\mathbf{^3}$University of Toronto and Vector Institute}\\
{\tt\small dvornik.nikita@gmail.com},
{\tt\small  garg@cs.toronto.edu}, \\
{\tt\small \{kosta, wildes\}@eecs.yorku.ca}, \\
{\tt\small \{isma.hadji, ran.zhang, allan.jepson\}@samsung.com} \\
}

\maketitle

\begin{abstract}
Instructional videos are an important resource to learn procedural tasks from human demonstrations. However, the instruction steps in such videos are typically short and sparse, with most of the video being irrelevant to the procedure. This motivates the need to temporally localize the instruction steps in such videos, i.e. the task called key-step localization. Traditional methods for key-step localization require video-level human annotations and thus do not scale to large datasets. In this work, we tackle the problem with no human supervision and introduce StepFormer, a self-supervised model that discovers and localizes instruction steps in a video.  StepFormer is a transformer decoder that attends to the video with learnable queries, and produces a sequence of slots capturing the key-steps in the video. We train our system on a large dataset of instructional videos, using their automatically-generated subtitles as the only source of supervision. In particular, we supervise our system with a sequence of text narrations using an order-aware loss function that filters out irrelevant phrases. We show that our model outperforms all previous unsupervised and weakly-supervised approaches on step detection and localization by a large margin on three challenging benchmarks. Moreover, our model demonstrates an emergent property to solve zero-shot multi-step localization and outperforms all relevant baselines at this task.

\end{abstract}

\section{Introduction}

\begin{figure}[t]
\centering
\includegraphics[trim=275 117 260 80,clip,width=0.49\textwidth]{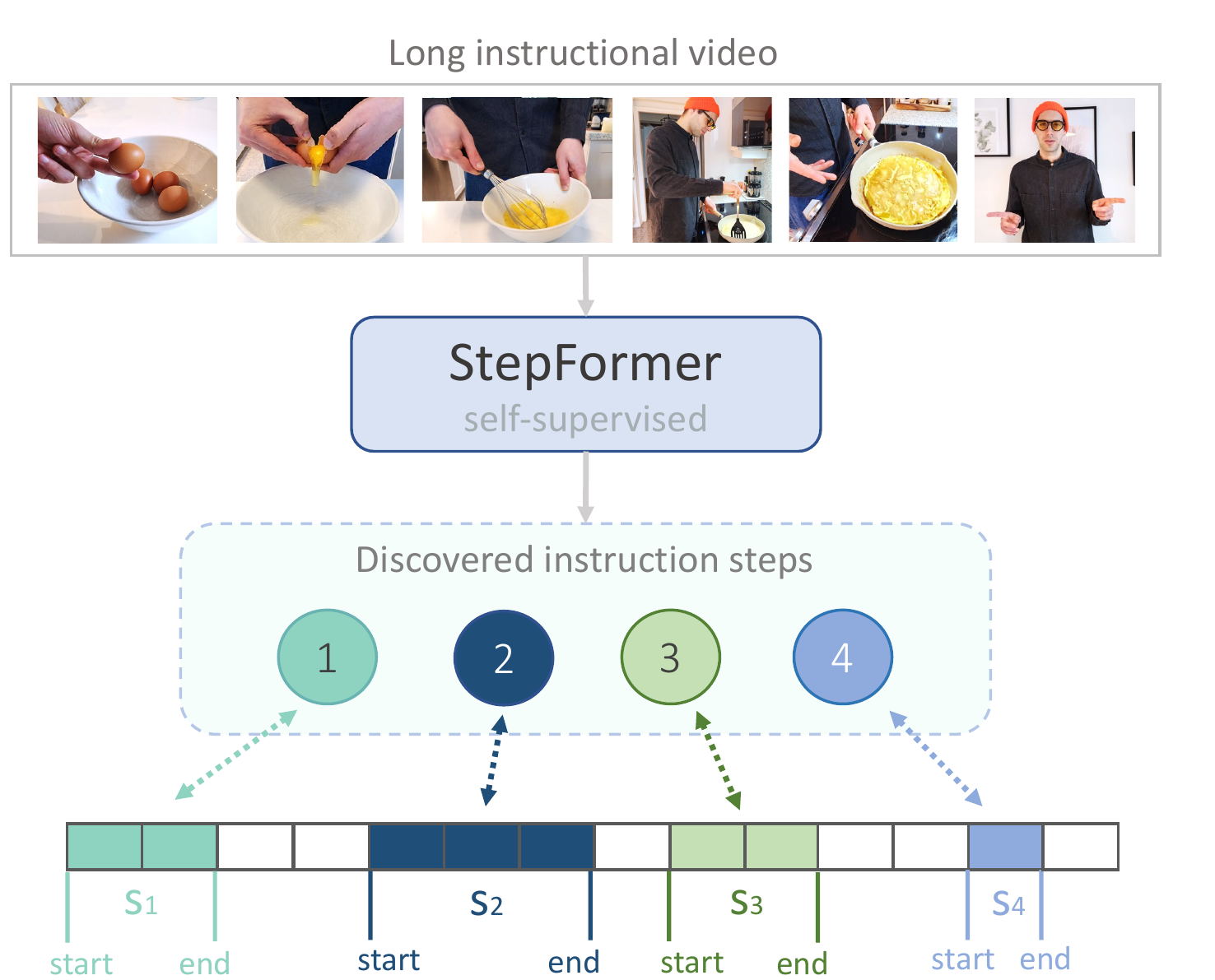}
\caption{\textbf{StepFormer for instruction step discovery and localization.}
StepFormer is a transformer decoder trained to discover instruction steps in a video, supervised purely from video subtitles.
At inference, it only needs the video to discover an ordered sequence of step slots and temporally localize them in the video.
\label{fig:teaser}
\vspace{-0.55cm}
}
\end{figure}

Observing someone perform a task (\eg cooking, assembling furniture or changing a tire) is a common approach for humans to acquire new skills.
Instructional videos provide an excellent large-scale resource to learn such procedural activities for both humans and AI agents. 
Consequently, instructional video datasets~\cite{CrossTask, COIN, miech19howto100m} have recently received significant attention and  been used for various video understanding tasks, \eg \cite{miech2020end,yang-hal-2021,procedure2020,bi2021procedure,p3iv,Sener2019,Girdhar2021}.
A potential impediment with using instructional videos from the web is that they tend to be long and noisy, \ie a limited number of
frames in the video correspond to the instruction steps,
while the remaining video segments are unrelated to the task (\eg title frames, close-ups of people talking and product advertisements). 
Thus, a major challenge 
with instructional videos is filtering out the uninformative frames and focusing only on the task-relevant segments, \ie the key-steps.
For instance, in a procedure of making a cake the key-steps could be ``crack eggs'', ``add sugar'', ``add flour'', then ``mix'', etc.
As a result, many recent efforts tackle the problem of instruction key-step localization, \eg \cite{CrossTask,COIN,miech19howto100m,miech2020end,dropdtw,graphdropdtw}.

Most previous work aiming at temporally localizing key-steps from instructional videos rely on some form of supervision.
Fully supervised approaches require start and end times of each step~\cite{COIN, YouCook2}.
Weakly supervised approaches either rely on knowledge of steps present in the video in the form of a set~\cite{lu2022set}, ordered steps transcript~\cite{dropdtw,CaoJCCN20} or partially ordered steps captured with a flow graph~\cite{graphdropdtw}.
Unsupervised approaches aim at directly detecting and localizing key-steps without a priori knowledge of instruction steps comprising videos~\cite{Shen2021,elhamifar2020self,kukleva2019unsupervised,TAN}.
Conceptually, these approaches are 
appealing for applications with 
large datasets, as they eschew the need for expensive and ambiguous labeling efforts.
In practice, previous unsupervised approaches 
rely on knowing the video-level task label at training time~\cite{Shen2021,elhamifar2020self,kukleva2019unsupervised}, and thus are not
fully unsupervised.
Moreover, so far they have been only applied to small instructional video datasets (up to 3K videos), as their task-specific models are not designed to handle large databases.
As a result, 
state-of-the-art procedure learning methods are not 
deployable at scale and without human supervision.

To address these challenges, we present StepFormer, a novel self-supervised approach that 
simultaneously discovers and temporally localizes procedure key-steps in long untrimmed videos, as illustrated in Figure~\ref{fig:teaser}.
StepFormer 
takes a video as input and outputs an ordered sequence of step slots, capturing instruction key-steps as they happen in the video.
Notably, it does not rely on any human annotations at training or inference time.  Instead, we train our model on a large instructional video dataset and use the accompanying narrations obtained from automated speech recognition (ASR)~\cite{Shen2021,tcn} as the only source of supervision.
StepFormer is implemented as a transformer decoder with learnable input queries.
Similar to the learnable object queries in DETR~\cite{DETR}, StepFormer's queries learn to attend to informative video segments and thus can be viewed as step proposals.
To enforce the output step slots to follow the temporal order, we use an order-aware loss based on temporal alignment of the learned steps and video narrations.
Since video narrations tend to be noisy and do not always describe 
visually groundable steps,
we use a flexible sequence-to-sequence alignment algorithm, Drop-DTW \cite{dropdtw}, which allows for non-alignable narrations to be dropped.
To localize the predicted step slots in the video, we explicitly use their learned temporal order.
Precisely, 
Drop-DTW 
aligns informative step slots with the video, and outputs start and end times for every detected step. 

We train our system on HowTo100M~\cite{miech19howto100m}, a large instructional video dataset with no human annotations.
For evaluation, we use three standard instructional videos 
benchmarks, \ie CrossTask~\cite{CrossTask}, ProceL~\cite{elhamifar2020self} and COIN\cite{COIN}. 
Empirically, for unsupervised step localization, our self-supervised method outperforms all weakly- and un-supervised baselines on all three downstream datasets, without any dataset-specific adaption.
Additionally, we demonstrate an emergent property of our model to perform zero-shot key-step localization from a text prompt (\ie without finetuning on the target dataset), where it also outperforms all relevant baselines.

\paragraph{Contributions.} The contributions of our paper are threefold.
\textbf{(i)} We present StepFormer, a novel self-supervised approach to key-step discovery and localization in instructional videos.
\textbf{(ii)} We model the temporal order of steps explicitly, and use it to design effective training and inference methods.
\textbf{(iii)} We supervise StepFormer only with video subtitles on a large instructional video dataset, and successfully transfer the model to three downstream datasets without any finetuning.
On all three datasets, StepFormer establishes a new state of the art on unsupervised step localization, outperforming unsupervised and weakly-supervised baselines.
We are commited to releasing our code.
\section{Related work}\label{sec:related_work}

\paragraph{Step localization in instructional videos.} 
Instruction step localization entails detecting and temporally localizing task-relevant video segments \cite{HuangFN16,DingX18,richard2018neuralnetworkviterbi,ChangHS0N19,CrossTask,miech2020end,Luo2020UniVL,dropdtw,graphdropdtw}. 
Work tackling this task can be largely split into three classes based on the type of supervision. Fully supervised approaches \cite{caba2015activitynet,MaFK16,COIN,YouCook2} are not ideal 
as they require expensive labelling efforts of key-step temporal endpoints in each video. Alternatively, weakly supervised approaches relax training efforts by either relying on unordered~\cite{lu2022set} or ordered step sequences~\cite{HuangFN16,DingX18,richard2018neuralnetworkviterbi,ChangHS0N19,CrossTask,dropdtw}. Nevertheless, relying on knowledge of steps still entails the labour intensive process of watching all videos to determine the steps present.
More recent work goes a step further to reduce labelling efforts by relying on task-level recipes \cite{graphdropdtw}.
Here, a general purpose recipe describing a given task is converted into a graph that captures the partial order of steps and is used to temporally localize the steps in videos.
However, this approach still requires human annotation for video task labels and creating the flow graphs.

More closely related to our work are unsupervised approaches that aim at discovering procedure step prototypes at training~\cite{elhamifar2020self,kukleva2019unsupervised}.
A common drawback of such methods is that they rely solely on the video signal to discover the step prototypes, completely ignoring the readily available complementary information present in the subtitles, which leads to subpar results.
More recent work trains for step localization by aligning video and narrations~ \cite{Shen2021}.
The important distinction from our work is that it does not explicitly tackle the target task of step detection and localization. Instead, their model only finetunes the video features via video to text alignment. 
To discover and localize the steps, they perform ad-hoc K-means clustering of the video features at test time.
In contrast, StepFormer is designed to discover individual instruction steps present in the video and localize them with an alignment procedure, which leverages the temporal structure of the process.
Crucially, all the above ``unsupervised'' methods actually rely on video-level task labels for guidance during training \cite{Shen2021,elhamifar2020self,kukleva2019unsupervised}; therefore, we refer to them as weakly-supervised in this paper.
Finally, the above methods can only be trained and evaluated on the same dataset, as the models they learn are task-specific.
In contrast, our StepFormer is a task-agnostic solution for step detection that can generalize across datasets; it only relies on videos and their accompanying narrations for training.
\vspace{-0.3cm}
\paragraph{Learning from visual-textual information.}
In the era of open world learning~\cite{joseph2021towards} and large scale datasets (\eg \cite{radford2021learning, miech19howto100m}), the ability to learn with minimal supervision is becoming vital.
For this reason, recent work strongly relies on the complementarity between visual and textual information as a natural source of supervision \cite{radford2021learning, miech2020end, Luo2020UniVL}.
This complementarity is especially appealing when dealing with narrated videos, such as instructional videos \cite{CrossTask,COIN,YouCook2}.
A large body of work uses this multimodal data for representation learning, thereby yielding strong representations enabling various downstream tasks~\cite{miech2020end, Luo2020UniVL, TAN}.
In this work, we build on these strong representations and further rely on weak temporal alignment between modalities to  directly tackle the problem of automatic key-step discovery and temporal localization.
\vspace{-0.3cm}
\paragraph{Sequence-to-sequence alignment.}
Alignment between sequences has seen a recent surge of interest with many approaches relying on it either as a proxy task for representation learning \cite{miech2020end,TAN,softdtw,d2tw,CaoJCCN20}, or more closely related to our work, for learning to localize steps \cite{dropdtw,ChangHS0N19,Shen2021,graphdropdtw}. 
Most previous work uses alignment in a weakly supervised setting, where knowledge of the sequence of steps is assumed \cite{ChangHS0N19,dropdtw,CaoJCCN20,Chang2021} or the video-level task label is used 
for clustering
\cite{Shen2021} or to build a task-level flow graph \cite{graphdropdtw}. In contrast, we do not make use of step order or video-label information. Instead, we train a model to discover steps conditioned on video content and align those discovered steps to narrations in a completely self-supervised manner. Notably, unlike other work using alignment \cite{dropdtw,graphdropdtw,TAN}, we devise a method that only requires both modalities at training time, while only requiring video at inference time to automatically discover and localize key steps.

\begin{figure*}[t]
	\centering
	\includegraphics[trim=15 300 35 35,clip,width=0.99\textwidth]{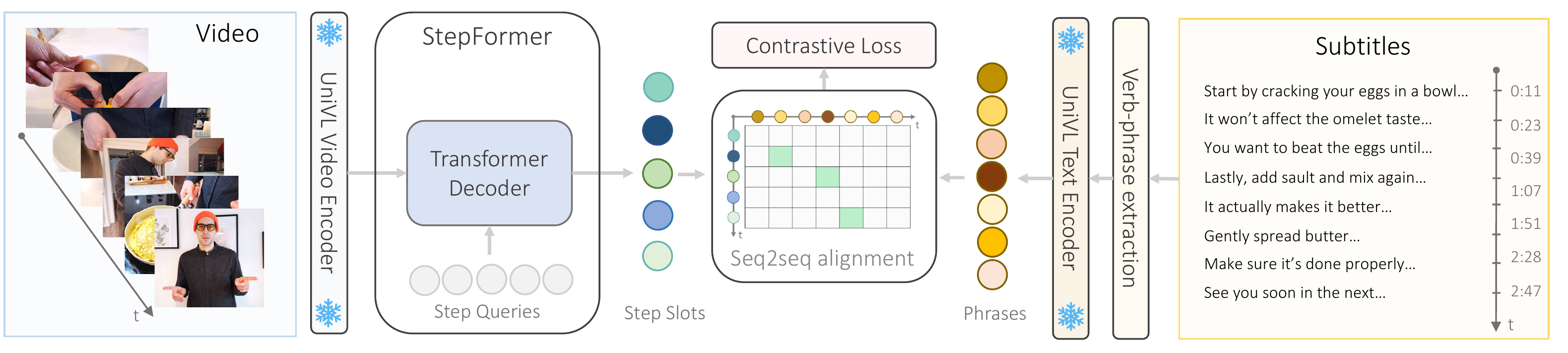}
	\caption{\textbf{StepFormer training overview.}
		(left) We first embed an untrimmed instructional video with a frozen UniVL encoder \cite{Luo2020UniVL}.  Next, we attend to the video with our StepFormer with learned step queries to extract step slots.
		(right) To form the training targets, we take the corresponding video subtitles, extract a sequence of verb phrases, and embed them with the UniVL text encoder \cite{Luo2020UniVL}.
		(middle) To supervise StepFormer, we find a matching subsequence between
		the step slots and
		verb phrases via seq-to-seq alignment with outlier rejection \cite{dropdtw}; 
		the green entries
		in the alignment matrix denote correspondences.
		The resulting alignment is used to define a contrastive loss.
		\label{fig:approach}
	}
 \vspace{-0.7cm}
\end{figure*}

\section{Technical approach}
In this section, we present our approach to self-supervised procedure step detection and localization in video. We first introduce StepFormer's architecture and our data preparation pipeline (Section~\ref{sec:stepformer}). Next, we describe the adopted  sequence-to-sequence alignment method (Section~\ref{sec:alignment}), which we use 
for both training and inference.
We then describe our training (Section~\ref{sec:training}) and inference (Section~\ref{sec:inference}) procedures.
Finally, we describe the implementation details in Section~\ref{sec:implementation}.

\subsection{StepFormer}\label{sec:stepformer}
StepFormer is our model for procedure step discovery in instructional videos. Given an $N$ second long video as input, StepFormer returns $K$ step slots, $\mathbf{s}$ -- a sequence of vectors capturing ordered instruction steps in the video.  We train StepFormer on a large dataset of instructional videos without any supervision by temporally aligning step slots with the narrations that accompany the video. Our full training pipeline is presented in Figure~\ref{fig:approach}.

\vspace{-0.3cm}
\paragraph{Data.}
To train StepFormer, we assume access to a large dataset of instructional videos, with \emph{no annotations}.
Our method is self-supervised; it relies on videos and their accompanying narrations. 
To use the narration for learning, we transform the speech into text using YouTube's ASR, and then follow the text processing pipeline proposed in previous work~\cite{Shen2021}.
More precisely, we run the subtitles through punctuation~\cite{Tilk2016BidirectionalRN} and co-reference resolution~\cite{spacy} modules, followed by a dependency parser to discover verb-phrases of the form \textit{verb+(prt)+dobj+(prep+pobj)}.
As a result, we transform long subtitle text into an ordered sequence of $L$ verb phrases, some of which describe groundable actions and procedure steps occurring in the video.
Examples of extracted verb phrases are given in supplement.

We extract features from video and phrases using UniVL, a self-supervised pre-trained video-language model~\cite{Luo2020UniVL}, such that verb phrases are mapped to a sequence of $L$ feature vectors, $\mathbf{p} \in \mathbb{R}^{L \times d}$, and an $N$ second video is mapped to a sequence of $N$ vectors, $\mathbf{v} \in \mathbb{R}^{N \times d}$.
Importantly, our approach relies on the fact that UniVL maps video clips and language into a shared embedding space, \eg a video of cutting a tomato and the phrase ``cutting a tomato'' map to similar vectors.
Starting with video and text features sharing a common embedding space makes training StepFormer feasible, despite the noise in the narrations.

While we derive supervision from the verb phrases, it is important to stress that in instructional videos, 
much of  
the audio narration contains irrelevant content, and only a small portion of the verb phrases are relevant to the instruction steps.
Thus, one of the challenges that we address in this work
is identifying the relevant phrases for training.
We describe how we leverage such noisy sequential targets to supervise our step slots in Section~\ref{sec:alignment}. 

\vspace{-0.3cm}
\paragraph{Architecture.}
We implement StepFormer as a multi-layer transformer decoder~\cite{transformer}, $\mathcal{T}$, that receives $K$ learnable queries, $\mathbf{q} \in \mathbb{R}^{K \times d}$, as input, and attends to the video features (with added sinusoidal positional embeddings~\cite{transformer}), $\mathbf{v} \in \mathbb{R}^{N \times d}$, at every decoder layer. Notably, while the length of the videos, $N$, can vary, 
the number of learned queries and corresponding step slots, $K$, is fixed.
The output of StepFormer is a sequence of $K$ contextualized vectors, $\mathbf{s} \in \mathbb{R}^{K \times d}$, that we term step slots, \ie $\mathbf{s} = \mathcal{T}(\mathbf{v}, \mathbf{q})$.
Intuitively, different step slots bind to different segments of the video; so, each step slot potentially represents a different instruction step. We enforce step slots to be temporally ordered, \ie a video segment captured by $\mathbf{s}_i$ would happen before the segment captured by $\mathbf{s}_j$, if $i < j$. 
To achieve the temporal ordering of step slots, we employ an order-aware sequence-to-sequence loss, described in Section~\ref{sec:training}.

\subsection{Sequence-to-sequence alignment}\label{sec:alignment}
A key feature of our StepFormer model is the \emph{temporal order} among the predicted step slots.
We take  full advantage of the temporal order in subtitles to supervise our system. 
Also, it enables us to localize the step slots in the video and respect their temporal order.
To infer the relationship between the elements of two sequences, we extensively use sequence-to-sequence alignment, which forms the basis of our method.

Aligning two sequences entails computing the optimal pairwise correspondence between the sequence elements, while preserving their match orderings.
Thus, given an ordered sequence of narrations for training,  sequence-to-sequence alignment is a natural choice to supervise the corresponding step slots and enforce their temporal order. 
Similarly, at inference, given a sequence of step slots and video frames that both follow the temporal order, it is natural to tackle step localization by aligning those sequences and finding correspondence between video frames and predicted step slots.
For sequence alignment, we choose the recent Drop-DTW~\cite{dropdtw} algorithm for its following properties.
\textbf{(i)} It operates on sequences of continuous vectors, such as video and text embeddings. \textbf{(ii)} It automatically detects outliers and allows to drop them from one or both sequences, essentially aligning only relevant sequence elements. \textbf{(iii)} It supports both one-to-one (needed for training) and many-to-one (needed for inference) matching.

In Drop-DTW's formulation, matching or dropping elements incurs some cost.
The cost of matching two elements typically is defined as their negative cosine similarity, while a drop cost is often defined as some percentile of the match cost distribution.
The alignment is then computed such that the total cost is minimized.
In this work, we are interested in the correspondence between elements, established by Drop-DTW through sequence alignment.
That is, given two vector sequences, $\mathbf{x} \in \mathbb{R}^{N \times d}$ and $\mathbf{z} \in \mathbb{R}^{K \times d}$, Drop-DTW returns a binary alignment matrix, $\mathsf{M}$ of size $K \times N$, indicating that elements $z_i$ and $x_j$ are matched if $\mathsf{M}_{ij} = 1$.
As described next, given the correspondence matrix, 
we can both formulate a self-supervised training objective and segment a video into steps at inference time.

\subsection{Training}\label{sec:training}
To train our system, we rely on a combination of losses, enforcing temporal alignment of step slots with narrations, and learning discriminative step slots, 
as well as additional training regularizers.
Below, we elaborate on each component of our training objective.


\vspace{-0.3cm}
\paragraph{Local step contrastive loss.}
We supervise the output of StepFormer, \ie the sequence of step slots, $\mathbf{s} \in \mathbb{R}^{K \times d}$, with the sequence of verb phrase embeddings, $\mathbf{p} \in \mathbb{R}^{L \times d}$. 
We first align the step slots, $\mathbf{s}$, with phrase embeddings, $\mathbf{p}$, using Drop-DTW.  This process allows verb phrases and step slots
that do not have a strong match to be dropped from further consideration,
and enforces only one-to-one correspondences between the elements that are a good match, \ie a single step slot can match with a maximum of one phrase, and vice versa.
The resulting correspondence matrix, $\mathsf{M}$, is used to construct positive and negative pairs, which we use in a contrastive training setting.
An example of such a correspondence matrix, $\mathsf{M}$, is given in Figure~\ref{fig:approach} (middle), with correspondences highlighted in green.
Specifically, every step slot, $\mathbf{s}_i$, and phrase embedding, $\mathbf{p}_j$, matched by Drop-DTW (\ie $\mathsf{M}_{ij} = 1$) forms a positive training pair, and all other non-matching pairs (\ie $\mathsf{M}_{ij} = 0$).
are used as negative examples.
To learn from such correspondences we use the Info-NCE~\cite{oord2018representation} loss, a contrastive objective that promotes the similarity between positive pairs, and pushes the negatives away from each other:
\begin{equation}
	\ell_{\text{NCE}}(\mathbf{s}_i, \mathbf{p}) = -\log \frac{f(\mathbf{s}_i, \mathbf{p}_{j^*})}{ f(\mathbf{s}_i, \mathbf{p}_{j^*}) + \sum_{j \ne j^*} f(\mathbf{s}_i, \mathbf{p}_j)},
\end{equation}
where $f(x,z) = \exp(\cos(x, z) / \gamma)$, $\gamma$ is a scaling temperature and $j^*$ is the index of $\mathbf{s}_i$'s positive pair.
The full sequence-to-sequence alignment loss, $\mathcal{L}_{\text{seq}}$, 
is defined as a combination of two Info-NCE losses, one contrasting the step slots
and the other contrasting the phrases:
\begin{equation}
	\mathcal{L}_{\text{seq}} = \frac{1}{K} \sum_{i=1}^{K} \ell_{\text{NCE}}(\mathbf{s}_i, \mathbf{p}) + \frac{1}{N} \sum_{j=1}^{L} \ell_{\text{NCE}}(\mathbf{p}_j, \mathbf{s}). \label{eq:loss_seq}
\end{equation}

\paragraph{Global video-step contrastive loss.}
The local step contrastive loss, (\ref{eq:loss_seq}), learns from positive and negative examples that all come from the same video;
however, it is known that contrastive learning can greatly benefit from a large and diverse set of negative examples~\cite{oord2018representation,sohn2016improved}.
Thus, we introduce an additional contrastive loss that forms negative pairs from phrases and step slots that come from different videos.
The intuition behind this loss is that, given a video, some of the extracted step slots must match with some phrases, regardless of the order, while at the same time, the slots and phrases coming from different videos should be different.

To realize the global contrastive loss, we use a contrastive multiple-instance-learning objective, MIL-NCE~\cite{miech2020end}, that promotes the similarity between sets of step slots and phrases coming from the same video, and contrasts them with the signal coming from different videos.
Formally, the global contrastive loss is defined as

\begin{equation}
	\mathcal{L}_{\text{glob}} = -\log \frac{1}{M} \sum_{i=1}^{M} \frac{\sum_{j \in \mathcal{P}_i} f(\mathbf{s}_i, \mathbf{p}_{j})}{\sum_{j \in \mathcal{P}_i} f(\mathbf{s}_i, \mathbf{p}_{j}) + \sum_{j \in \mathcal{N}_i} f(\mathbf{s}_i, \mathbf{p}_j)},
\end{equation}
where $f(x,z) = \exp(\cos(x, z) / \gamma)$, $\gamma$ is a scaling temperature, $M = K \cdot B$ is the total number of step slots across the batch of size $B$, $\mathcal{P}_i$ is the set of indexes $j$ that form a positive pair with $\mathbf{s}_i$, \ie the $\mathbf{p}_j$ coming from the same video, and $\mathcal{N}_j$ indexes the negative pairs, coming from different videos.
An additional benefit of this global loss is that step slots representing non-procedural steps, commonly appearing across different videos, will be discouraged.
As we show in the experiments, this loss greatly improves the learned step slots.

\paragraph{Regularization.}
To improve the training and introduce prior knowledge in the system, we use two additional regularizers acting on the step slots. 

The first regularizer enforces diversity among step slots.
Indeed, the main objective functions described so far do not encourage the step slots predicted from one video to be significantly different, which results in numerous duplicate steps capturing the same part of the video.
To counter this limitation, we introduce a regularization term, $\mathcal{L}_{\text{div}}$, that encourages low cosine similarity among step slots of the same video and thereby make predicted step slots more diverse.

The second regularizer enforces attention smoothness.
Due to natural video continuity, we expect the attention of step slots in the video to change smoothly, and be similar for close frames.
Thus, we add a regularization term, $\mathcal{L}_{\text{smooth}}$, that, for all step slots, enforces their attention in the video to be similar for neighboring frames and different for distant frames.
Both regularizers improve discovered step and segmentation quality, as we show in the experiments.
Please see supplement for the exact loss definitions.

Finally, to train our system we optimize the weighted sum of four losses,
\begin{equation}
	\mathcal{L}_{\text{total}} = \mathcal{L}_{\text{seq}} + \mathcal{L}_{\text{global}} + \alpha \mathcal{L}_{\text{div}} + \beta \mathcal{L}_{\text{smooth}},
\end{equation}
where the $\alpha$ and $\beta$ scalars are chosen using cross validation.

\begin{figure}[t]
	\centering
	\includegraphics[trim=245 120 175 115,clip,width=0.49\textwidth]{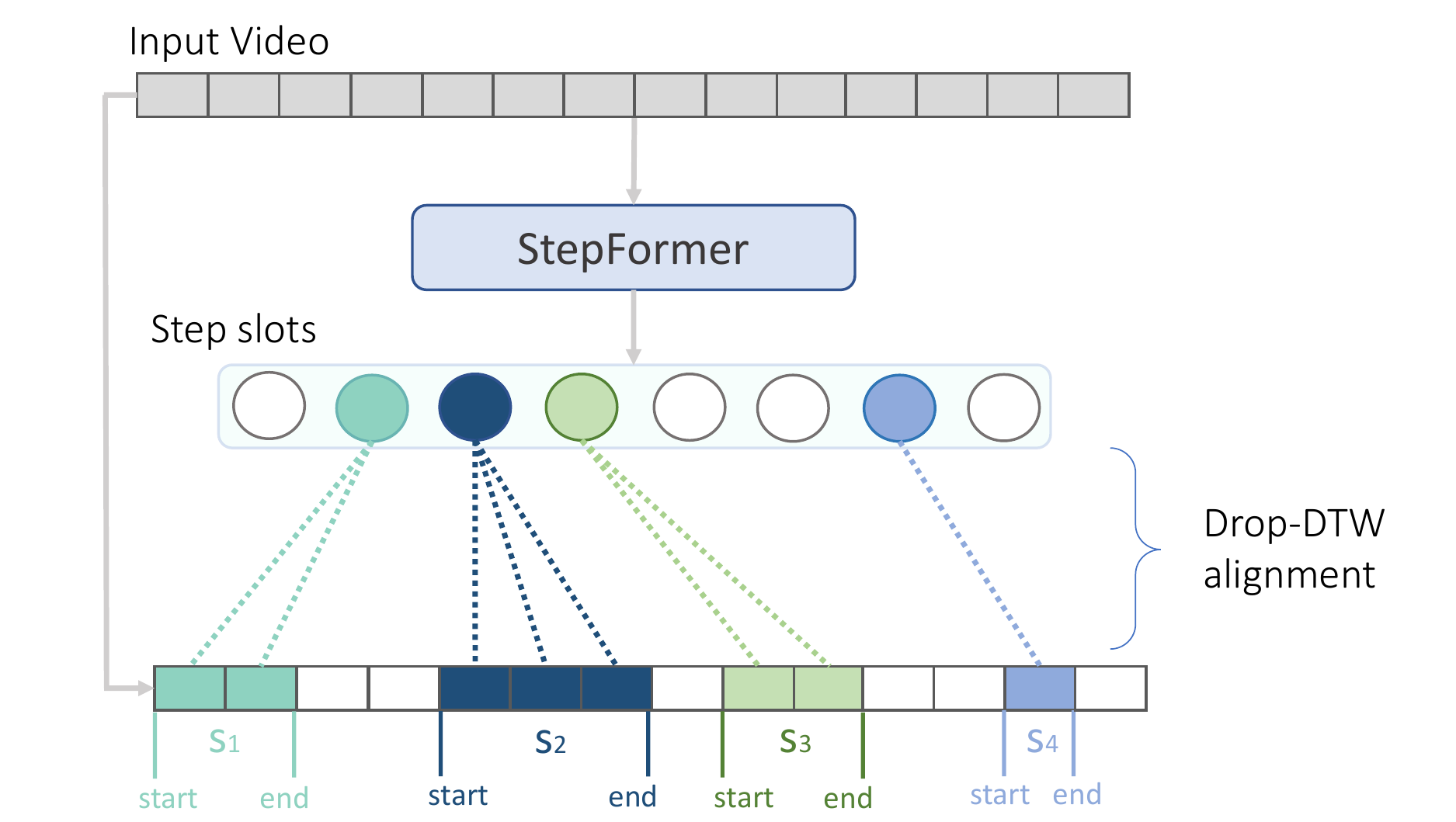}
	\caption{
	    \textbf{Step localization inference.}
		Given a video, StepFormer first extracts a sequence of step slots.
		The step slots can be interpreted
		as temporally ordered key-step candidates.
		Next, it aligns the sequence of step slots with the video sequence, using Drop-DTW~\cite{dropdtw}.  This step puts the elements of the two sequences in correspondence, while identifying outliers.
		Here, the step slots and video frames of the same color are matched by Drop-DTW, while the white slots and frames are dropped from the alignment.
		The colored video segments represent the final step localization result.
		\label{fig:inference}
	}
 \vspace{-0.6cm}
\end{figure}

\subsection{Step localization inference}\label{sec:inference}
At inference time, StepFormer detects and localizes key-steps without requiring narrations.
StepFormer's inference procedure is illustrated in Figure~\ref{fig:inference}.
It takes an instructional video as input and returns an ordered sequence of $K$ step slots conditioned on this video.
Each step slot corresponds to a procedural key-segment in the video and carries its semantics.
However, given that we use a large fixed number of $K$ step slots for all videos, some step slots may be duplicates or have weak binding with the video segment.
For this reason, we need select a subset of the step slots that concisely describe the given video. 
A simple approach would be to use hard attention of step slots in the video directly~\cite{locatello2020object,Shen2021} and get some meaningful step segments.
While viable, this solution can not properly handle the duplicates, and more importantly, it completely ignores the order present in step slots, which is a useful source of information for step localization~\cite{dropdtw,ChangHS0N19}.

Instead, we rely on sequence-to-sequence alignment as our inference procedure for step localization.
Specifically, for each input video, we first extract the corresponding $K$ step slots using StepFormer.
We then use Drop-DTW~\cite{dropdtw} to align the step slot embeddings, $\mathbf{s}$, to the video embeddings, $\mathbf{v}$, allowing outliers to be dropped from both sequences. Notably,
here, we use the formulation of Drop-DTW that allows many-to-one correspondences, so it can assign multiple video frames to a single step slot, which effectively segments the video into steps. 

The ability to drop irrelevant step slots for individual videos, and the ability to take the order of steps into account explicitly leads to improved segmentation quality, as we later validate empirically.
Notably, alignment of step slots and video is possible in the first place, because the video and text features used for training share a common embedding space. Thus, step slots, which were trained using procedure text should align with video too.

\subsection{Implementation details}\label{sec:implementation}
We implement StepFormer using a Pre-LN Transformer decoder~\cite{xiong2020layer} with six layers, and 
fix the number of step slots to be $K=32$.
We train the model for a total of 60 epochs, starting with three epochs of warm-up from 0 to 3e-4, followed by cosine decay to 1e-6.
We use weight decay of 1e-4 and 0.1 dropout~\cite{srivastava2014dropout} in the intermediate layers.
In our sequence-to-sequence alignment loss, we use Drop-DTW with 0.8 percentile drop cost~\cite{dropdtw}.
We set the regularizer weights to $\alpha=0.3$ and $\beta=0.02$.
All the hyper-parameters are set using cross-validation.
It takes about two days to train one model using eight Tesla T4 GPUs. 

\section{Experiments}

\begin{table*}[t]
\centering
\resizebox{0.9\textwidth}{!}{
	\begin{tabular}{l|l|llll|llll|llll}
	\hline
	\multirow{2}{*}{Supervision} & \multirow{2}{*}{Method}	& \multicolumn{4}{c|}{CrossTask} &  \multicolumn{4}{c|}{ProceL}  & \multicolumn{4}{c}{COIN}   \\
	  & 	& F1 & Prec.\ & Rec. & MoF &  F1 & Prec.\ & Rec. & MoF  & F1 & Prec.\ & Rec.\ & MoF   \\ 
		\hline
      \multirow{3}{*}{Weak}      & Kukleva et al.~\cite{kukleva2019unsupervised} & 15.9 & 10.3 & 37.2 & 28.5 & 19.4 & 14.1 & \textbf{34.3} & 30.4 & 28.5 & 21.5 & 48.6 & 8.2 \\
    & Elhamifar et al.~\cite{elhamifar2020self} & 17.2 & 10.9 & 41.7 & 41.7 & 15.5 & 11.3 & 27.3 & 27.3 & 28.4 & 20.6 & \bf{52.0} & 30.2 \\ 
	&	Shen et al.~\cite{Shen2021} & 22.3 & 16.2 & 37.5 & 41.0 & 20.3 & 17.4 & 26.7 & 35.4 & 30.1 & 25.9 & 39.8 & 35.6  \\ 
		\hline
	\multirow{3}{*}{Unsupervised}& Kukleva et al.~\cite{kukleva2019unsupervised} & 18.1 & 12.0 & 39.8 & 29.4 & 17.2 & 12.8 & 30.0 & 28.9 & 27.9 & 21.8 & 45.0 & 13.6    \\
	& Shen et al.~\cite{Shen2021}	& 16.0 & 11.7 & 23.1 & 37.1 & 18.0 & 13.6 & 22.8 & 33.1 & 18.8 & 16.5 & 25.1 & 32.7  \\
	& StepFormer (ours)	& \textbf{28.3} & \textbf{22.1} & \textbf{42.0} & \textbf{41.9} & \textbf{21.9} & \textbf{18.3} & {28.1} & \textbf{40.9} & \textbf{32.1} & \textbf{27.1} & {42.5} & \textbf{38.6} \\
	\hline
	\end{tabular}
	}
	\caption{Dataset-level unsupervised step detection and localization results on the CrossTask, ProceL and COIN datasets.} \label{tab:baseline_comparisons}
    \vspace{-0.2cm}
\end{table*}

In this section, we evaluate StepFormer on unsupervised step detection and localization (Section~\ref{sec:baselines}), and its emergent property to address zero-shot step localization (Section~\ref{sec:zeroshot}).
Finally, we perform an ablation study of our approach (Section~\ref{sec:ablation}).

\subsection{Evaluation protocol} \label{sec:exp-protocol}
\noindent {\bf Datasets.} Given that our approach is trained entirely in a self-supervised manner, we train our model on the largest instructional video dataset, HowTo100M \cite{miech19howto100m}, that comes with no annotations.
Following previous work \cite{Shen2021}, we evaluate our proposed method on the CrossTask \cite{CrossTask} and ProceL \cite{elhamifar2020self} datasets.
CrossTask contains 2750 videos, depicting 18 different tasks, while ProceL covers 12 different tasks with 60 videos for each task.
In addition, given that our approach does not require any finetuning, we also report results on the largest annotated instructional video dataset, \ie COIN \cite{COIN}, which includes 11827 videos, covering 778 procedures.
For all evaluations, the same model pre-trained on HowTo100M is used directly, without finetuning or adaptation to the downstream task.
\vspace{-0.2cm}
\paragraph{Evaluation metrics.}
Following previous work \cite{Shen2021}, we use precision, recall, F1 score and Mean over Frames (MoF) for evaluation. 
Precision highlights the percentage of frames correctly predicted as belonging to key-steps, while recall emphasizes the model's ability to detect frames belonging to key-steps. F1 score balances the two previous metrics by calculating their harmonic mean and is therefore the most informative metric. MoF is a less strict metric calculating the percentage of correctly predicted frames, including the background ones (\ie non key-steps).
\vspace{-0.2cm}
\paragraph{Baselines.}
We compare our method to previous weakly-supervised baselines~\cite{kukleva2019unsupervised,elhamifar2020self,Shen2021}, which only use the video task label as supervision.
Precisely, Kukleva et al.~\cite{kukleva2019unsupervised} and Elhamifar et al.~\cite{elhamifar2020self} train a model purely from video, to discover global step prototypes in the training set, while Shen et al.~\cite{Shen2021} extract prototypes from a video on the fly and supervises them with text subtitles.
In this work, we adapt the methods of Kukleva et al.\ and Shen et al.\ to completely unsupervised training (\ie without video labels), and perform a more direct comparison to our unsupervised StepFormer.
All the baseline methods must be trained and tested on the same dataset, and cannot generalize to new data, unlike StepFormer.
The supplement provides additional details about our baselines.

\subsection{Unsupervised step discovery and localization}\label{sec:baselines}
The goal of this experiment is to evaluate our model's ability to automatically discover task-relevant steps and localize them.
Thus, the model does not have access to ground-truth key-steps in this experiment.
Instead, for each video, we predict $K$ step slots with our StepFormer, and let our inference procedure select only the task-relevant steps among them (see Section~\ref{sec:inference}).
To do so, we temporally align a video with its corresponding step slots while allowing both ``outlier'' step slots and frames to be dropped.
To achieve this end, we use Drop-DTW \cite{dropdtw}, which performs temporal sequence alignment while dropping interspersed outliers in both sequences.
The output of this alignment is a sequence of $K'$ step slots ($K' \le K$) and corresponding video segments, which our model highlights as corresponding to key, \emph{task-relevant}, steps. 

To evaluate the quality of the extracted key-steps, we follow previous work \cite{Shen2021,elhamifar2020self,kukleva2019unsupervised} to associate predicted steps to ground-truth segments using clustering and Hungarian matching. Specifically, we start by accumulating the extracted key-steps of all videos corresponding to a given task (\eg making pancake, changing a car tire, etc.).
Next, we cluster all extracted steps using K-means, setting the number of clusters, $K_c$, to be the ground-truth number of key-steps in the task according to the dataset.
In each cluster, we only keep $60\%$ of the step slots that are the closest to the cluster center and label the rest of the steps as background, as done in previous work~\cite{Shen2021}.
Finally, we use Hungarian matching to find a global one-to-one matching between the video segment associated with each cluster and the ground-truth segments. We evaluate the resulting corresponding segments using the metrics defined in Section~\ref{sec:exp-protocol}.

\begin{figure}[t]
\centering
\includegraphics[trim=207 222 170 110,clip,width=0.49\textwidth]{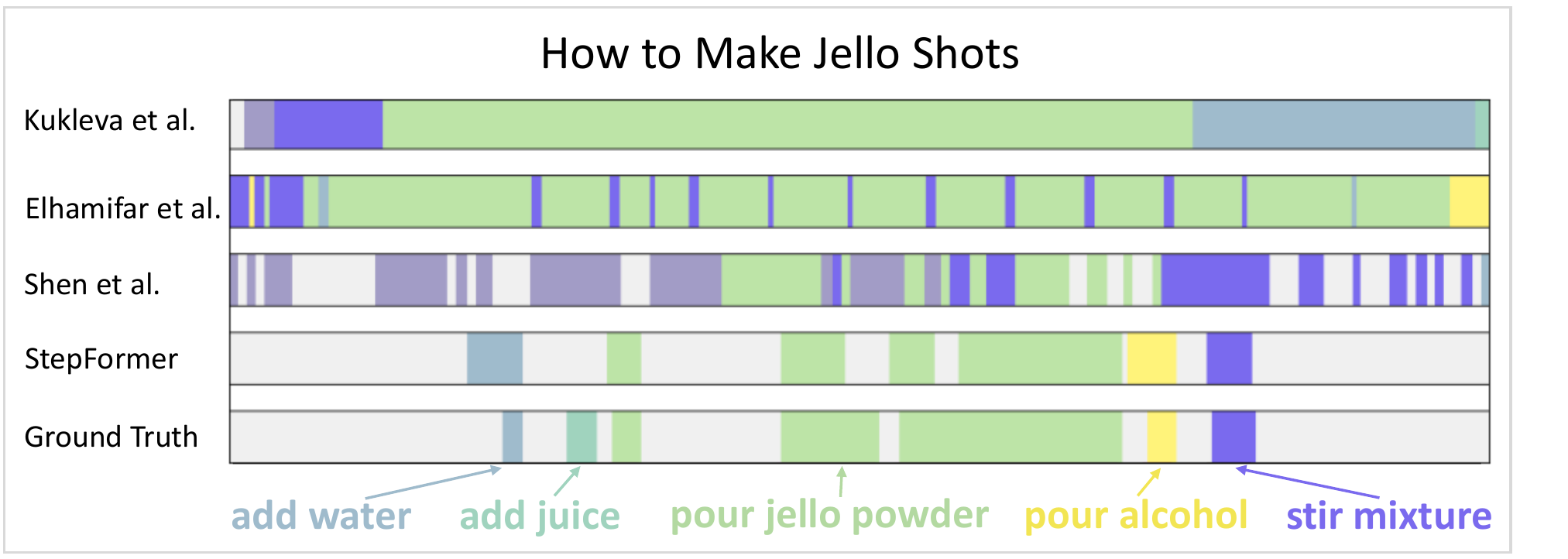}
\caption{\textbf{Qualitative comparison of temporal localization.}
Comparison of our self-supervised StepFormer 
with the weakly-supervised baselines~\cite{kukleva2019unsupervised,elhamifar2020self,Shen2021} on CrossTask.
\label{fig:segmentation}
}
\vspace{-10pt}
\end{figure}

Table~\ref{tab:baseline_comparisons} compares our performance to the baselines~\cite{kukleva2019unsupervised,elhamifar2020self,Shen2021}, which all use a video task label for training. 
We report baseline results under their original settings, where they use video task label for training, which we refer to as weak supervision.
Additionally, we compare to \cite{kukleva2019unsupervised,Shen2021}
adapted to the fully unsupervised setting (\ie we train a single model for all tasks).
The results summarized in Table~\ref{tab:baseline_comparisons} speak decisively in favor of our approach, which outperforms all baselines in both the weak and unsupervised settings with sizeable margins.  This pattern is despite the fact that the baselines are trained and tested on the same data distribution, whereas our model generalizes to the target dataset without any finetuning.
The results \textbf{(i)} highlight StepFormer's capability to automatically discover key-steps even while being completely unsupervised and having no access to the datasets used for evaluation during training and \textbf{(ii)} also show that our method yields better localization, as highlighted in the qualitative results in Figure~\ref{fig:segmentation}.

\begin{table*}[t]
\centering
\resizebox{0.9\textwidth}{!}{
	\begin{tabular}{l| l|llll | llll | llll}
	\hline
 
	\multirow{2}{*}{Supervision} & \multirow{2}{*}{Method}	& \multicolumn{4}{c|}{CrossTask} &  \multicolumn{4}{c|}{ProceL}  & \multicolumn{4}{c}{COIN}   \\
	&	& IoU & Prec. & Rec. & MoF & IoU & Prec. & Rec. & MoF & IoU & Prec. & Rec. & MoF  \\ 
		\hline
	Weak & Shen et al.~\cite{Shen2021}	& 16.3 & 27.2 & 26.2 &  67.2 & 12.1 & 16.0 & 19.0 & 30.7 & 20.8 & 31.3 & 33.7 & 40.5 \\
    \hline
	\multirow{3}{*}{Unsupervised} & GT text features	& 18.6 & 32.2 & 29.2 &  \textbf{67.9} & 12.6 & 20.2 & 21.1 & 36.1 & 24.4 & 38.1 & 35.3 & \textbf{50.6} \\
	& Shen et al.~\cite{Shen2021}	& 12.8 & 25.2 & 19.9 &  66.9 & 10.5 & 13.8 & 14.7 & 24.2 & 11.7 & 18.3 & 24.3 & 29.2 \\
	& Ours     	& \textbf{23.7} & \textbf{32.9} & \textbf{43.1} &  67.1 & \textbf{18.1} & \textbf{36.3} & \textbf{24.8} & \textbf{36.9} & \textbf{27.1} & \textbf{42.8} & \textbf{37.9} & 48.4 \\
		\hline
	\end{tabular}
	}
	\caption{Zero-shot key-step localization on the the CrossTask, ProceL and COIN dataset. GT denotes ground-truth.} \label{tab:step-filtering-with-gt}
	\vspace{-10pt}
\end{table*}

\begin{figure*}[t]
\centering
\includegraphics[trim=0 345 0 30,clip,width=0.99\textwidth]{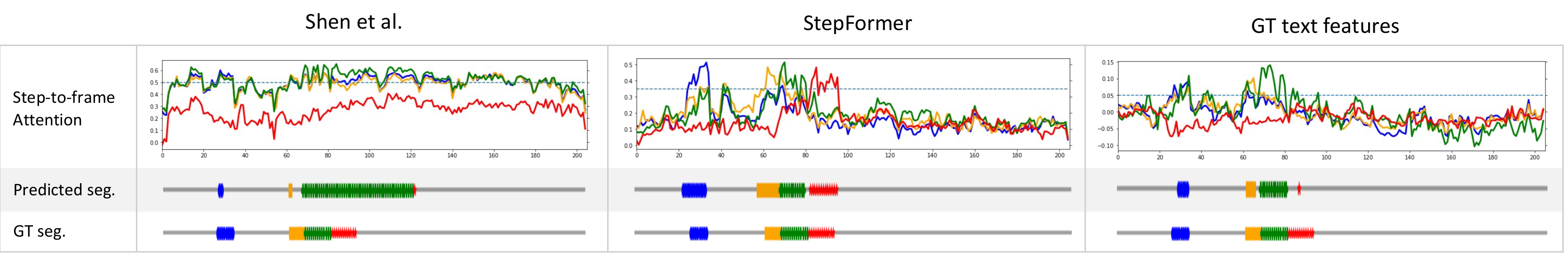}
\caption{\textbf{Visualization of step attention in the video for zero-shot multi-step localization.}
(top row) The attention of the step slots in the video, with different colors capturing different steps.
(middle row) Multi-step localization results obtained from the above attention.
(bottom row) The corresponding ground-truth step localization.
Notably, the step slot attention (column 3) even outperforms the attention of GT text descriptions (column 4) at step localisation, without ever being trained to do so.
\label{fig:attention}
}
\vspace{-5pt}
\end{figure*}

\subsection{Zero-shot key-step localization}\label{sec:zeroshot}
In this experiment, the goal is to show that the step slots learned by our StepFormer can be used directly for key-step localization in a zero-shot setting.
The task of zero-shot step localization is defined as temporally localizing a sequence of procedure steps in the video, given their natural text description (\eg ``break egg'', ``add sugar'', ``add flour'', etc.) without any training or finetuning on the target data.
To tackle this task, we use each video in the target dataset as input to our StepFormer and extract the corresponding $K$ step slots.
These slots are then aligned to the embeddings of the ground-truth steps. 
We use Drop-DTW \cite{dropdtw} to compute an alignment between the $K$ step slots and ground-truth steps while allowing unmatched slots to be dropped.  
Next, we align the remaining step slots, corresponding to ground-truth steps, with the video (again using Drop-DTW) to automatically identify the start and end times of each ground-truth step.
Notably, to address this task successfully,
the step slots must align well with both text and video features simultaneously. 
Thus, we use this evaluation as an opportunity for a deeper investigation on our model's capability.

Table~\ref{tab:step-filtering-with-gt} summarizes our results under these settings, while comparing to a state-of-the-art method for step detection and localization \cite{Shen2021} in both weakly supervised (\ie separate model for each task) and unsupervised (a single model for all tasks) settings.
Note that out of all available baselines, we compare only to Shen et al.~\cite{Shen2021}, as it is the only method extracting video clusters alignable with text using subtitle text supervision.
The results show the superiority of our method and highlight the quality of the learned step slots, which are better able to capture procedure steps given the sequence of ground-truth text descriptions.

We include an additional baseline, dubbed ``GT text features'', where we directly use embedded ground-truth step text to localize the key-steps in the video, using Drop-DTW to align the text and video features.
This is a strong baseline as it relies on aligning UniVL video and text features, that were trained to be alignable in the first place.
Interestingly, StepFormer outperforms this baseline as well.
We attribute this performance improvement to two key aspects of our model.
First, StepFormer is trained to align \emph{entire videos} with their subtitles, which allows StepFormer to globally reason about the step order, giving it an advantage over the original UniVL features, extracted locally.
Second, step slots are directly conditioned on video and therefore have higher potential of better aligning with video content.

To better understand the behavior of our step slots and highlight their usefulness, in Figure~\ref{fig:attention}, we demonstrate the attention of the selected step slots (or ground-truth text features) in the video.
While Shen et al.\ \cite{Shen2021} outputs steps unalignable with the video, StepFormer's step slots accurately capture independent procedure steps and provide clear attention peaks in the true key-step locations.
Further, the quality of the StepFormer outputs even surpasses those of the ground-truth text features.

\begin{table}[t]
\centering
\resizebox{0.99\columnwidth}{!}{
	\begin{tabular}{l|llll|llll}
	\hline
	\multirow{2}{*}{Method} & \multicolumn{4}{c|}{Unsup. Segmentation} & \multicolumn{4}{c}{Zero-shot Localization}  \\
		& F1 & Prec. & Rec. & MoF  & IoU & Prec. & Rec. & MoF \\ 
		\hline
	Ours w/o $\mathcal{L}_{\text{seq}}$	& 17.3 & 18.4 & 20.5 &  \textbf{63.5} & 4.6 & 5.9 & 6.3 & \textbf{69.8} \\
	Ours w/o $\mathcal{L}_{\text{global}}$	& 20.9 & 14.9 & 41.1 &  33.3  &11.5 & 17.9 & 26.2 & 55.2 \\
	Ours w/o $\mathcal{L}_{\text{div}}$	& 26.7 & 20.5 & 40.9 &  44.6 & 16.9 & 27.6 & 29.3 & 66.1 \\
	Ours w/o $\mathcal{L}_{\text{smooth}}$	& 27.1 & 21.2 & 40.0 &  48.7  & 17.7 & 26.9 & 33.0 & 63.4 \\
	Ours w/o Drop-DTW & 28.1 & 20.3 & 41.4 &  47.0  & 20.1 & 28.0 & 40.8 & 62.1 \\
		\hline
    Ours	& \textbf{28.3} & \textbf{22.1} & \textbf{42.0} & 41.9 & \textbf{23.7} & \textbf{32.9} & \textbf{43.1} & 67.1 \\
		\hline
	\end{tabular}
	}
	\caption{Ablation study of StepFormer's training and inference components on CrossTask.} \label{tab:ablation}
	\vspace{-15pt}
\end{table}

\subsection{Ablation Study}\label{sec:ablation}
In this section, we perform an ablation study of our approach by systematically removing StepFormer's components and measuring the performance on unsupervised step segmentation and zero-shot key-step localization on CrossTask. 
An additional study on the hyper-parameters can be found in the supplement.
The ablation results are summarized in Table~\ref{tab:ablation}.
Removing the main sequence-to-sequence alignment loss, $\mathcal{L}_{\text{seq}}$, severely degrades the performance of our model, especially on zero-shot key-step localization, where step order is essential.
The global contrastive learning loss, $\mathcal{L}_{\text{global}}$, is also important for learning expressive step slots, removing it impairs the performance on both tasks.
Finally, removing the diversity, $\mathcal{L}_{\text{div}}$, and smoothness, $\mathcal{L}_{\text{smooth}}$, regularizers of our step slots mildly hurts unsupervised step segmentation performance.
However, the drop in performance of zero-shot key-step localization is more noticable, as accurate key-step localization requires step slots to be diverse (when aligning them to step text descriptions), and have smooth attention in the video (when aligning the slots to the video).
Finally, we replace our step localization inference procedure based on Drop-DTW, with an order-agnostic procedure, that assigns a frame to the most similar step-slot, akin to~\cite{Shen2021,elhamifar2020self}.
This modification does not affect the performance on unsupervised segmentation significantly, because the task evaluation is insensitive to the step order.
However, there is a notable drop in performance on zero-shot key-step localization, where recovering the correct step order is essential.
Overall, all the proposed components of StepFormer are important to achieve state-of-the-art step discovery and localization.
\section{Conclusion}
We introduced StepFormer, a self-supervised, 
task-agnostic model that discovers and localizes key-steps in instructional videos.
We train it on a large dataset of instructional videos using automatically generated subtitles as the only source of supervision.
Our model yields state-of-the-art results on step localization across multiple datasets, even outperforming weakly-supervised models.
StepFormer not only provides good step localization, it defines a new type of model for procedure understanding, that is simple, effective, scalable and requires no supervision. Therefore, it has potential to 
advance research in 
procedure understanding.
\newpage
\appendix
\begin{center}  
\huge{Appendix}
\end{center}
\section{Summary}
In this supplemental material, we elaborate on the details of the proposed approach and experimental validation setup. In Section~\ref{sec:subtitles}, we begin by supplementing the description of data processing provided in Section 3.1 of the main paper with an explicit illustration of the subtitle processing outcome. Next, we define the regularization losses in Section~\ref{sec:reg}. Then, in Section~\ref{sec:baselines}, we give a more detailed description of the used baselines. Finally, we present the additional ablation studies in Section~\ref{sec:ablation}.

\section{Subtitle processing}\label{sec:subtitles}
\begin{figure*}[t]
	\centering
	\includegraphics[trim=5 50 45 35,clip,width=0.99\textwidth]{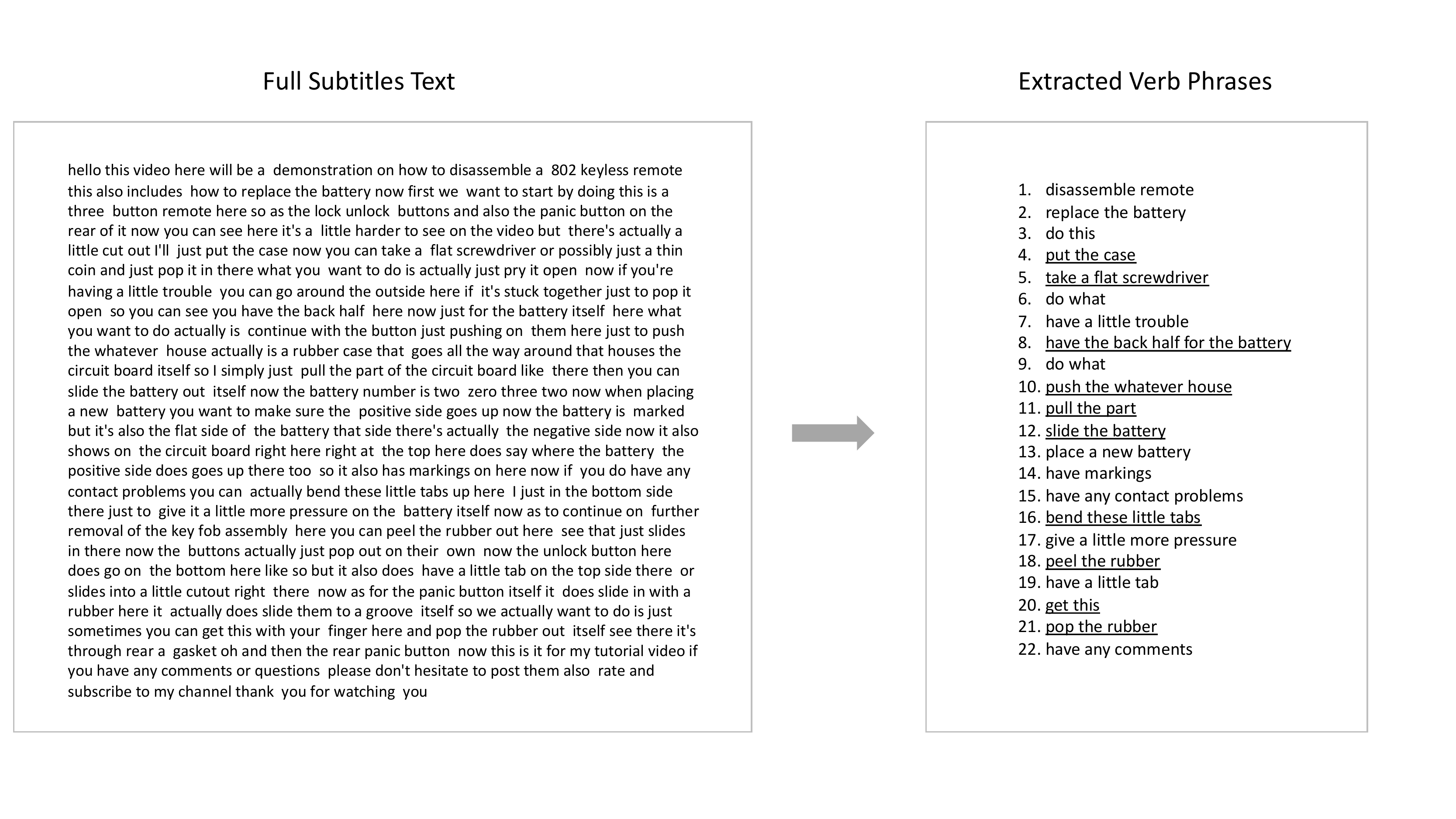}
	\caption{\textbf{Verb-phrase extraction from subtitles.}
		(left) Original video subtitles.
		(right) Verb phrases extracted from the subtitles. The underlined verb phrases are the ones chosen by Drop-DTW for step slot supervision at end of training.
		\label{fig:phrases}
	}
 \vspace{-0.3cm}
\end{figure*}

To train StepFormer, we derive supervision from video subtitles (or narrations), as described in Section 3.1 of the main paper.
More precisely, we run the narrations through punctuation~\cite{Tilk2016BidirectionalRN} and co-reference resolution~\cite{spacy} modules, followed by a dependency parser to discover verb-phrases of the form \textit{verb+(prt)+dobj+(prep+pobj)}.
As a result, we transform long subtitle text into an ordered sequence of verb phrases, some of which describe groundable actions and procedure steps occurring in the video.
We demonstrate an example of the verb phrase extraction in Figure~\ref{fig:phrases}, and additionally highlight the phrases that were matched to step slots during training (in the last epoch).
This example, confirms that a subset of the extracted verb phrases indeed contain important information about procedure steps. Notably, most of the relevant verb phrases get selected by Drop-DTW for supervision, as shown in the underlined steps on the right of Figure~\ref{fig:phrases}.

\section{Regularization losses}\label{sec:reg}
As described in Section~3.3 of the main paper, we train StepFormer with verb phrases supervision and employ two extra regularizers, acting on the step slots.

\paragraph{Diversity regularizer.} 
The first regularizer, $\mathcal{L}_{\text{div}}$, enforces diversity among step slots.
Precisely, given the step slots $\mathbf{s} \in \mathbb{R}^{K \times d}$ extracted from video $\mathbf{v} \in \mathbb{R}^{N \times d}$ using the transformer, $\mathcal{T}$, i.e., $\mathbf{s} = \mathcal{T}(\mathbf{v})$, the diversity regularizer encourages low cosine similarity among the step slots $\mathbf{s}_i$ as follows:
\begin{equation}
\mathcal{L}_{\text{div}} = \frac{1}{K (K - 1)} \sum_{i=1}^{K}\sum_{j \neq i} \cos(\mathbf{s}_i, \mathbf{s}_j),
\end{equation}
Where $K$ is the number of predicted slots.
The diversity regularizer promotes slot diversity and improves performance by removing duplicate slots as validated in Table~3 of the main paper.

\paragraph{Smoothness regularizer.} 
The second regularizer, $\mathcal{L}_{\text{smooth}}$, enforces that step slots attend to video content smoothly. 
Intuitively, due to natural video continuity, we expect the attention of step slots in the video to change smoothly, and be similar for close frames.
Given the video, $\mathbf{v} \in \mathbb{R}^{N \times d}$, and the corresponding step slots, $\mathbf{s} \in \mathbb{R}^{K \times d}$, the attention, $\mathbf{a} \in \mathbb{R}^{N \times K}$, of all steps in the video is defined as follows:
\begin{equation}
\mathbf{a} = \text{softmax}(\cos(\mathbf{v}, \mathbf{s}) / \gamma),
\end{equation}
where $\gamma = 0.03$ is the softmax temperature, softmax is taken along the last dimension, and $\cos(\mathbf{v}, \mathbf{s}) \in \mathbb{R}^{N \times K}$ is a matrix of cosine similarities, \ie $\cos(\mathbf{v}, \mathbf{s})_{ij} = \cos(\mathbf{v}_i, \mathbf{s}_j)$.

To define the regularizer on the attention vectors, we draw inspiration from the video representation learning literature~\cite{liang2022self}.
In particular, we require that, for a given attention vector $\mathbf{a}_i \in \mathbb{R}^K$ (at frame $i$), the attention vectors $\mathbf{a}_k$ of the neighboring frames, \ie $\|i - k \| \le M$, are similar, and the attention vectors, $\mathbf{a}_j$ of distant frames, \ie $\|i - j \| > M$, are dissimilar, where $M$ is the size of the positive neighborhood.
We implement this requirement using the MIL-NCE contrastive loss on the attention according to

\begin{equation}
	\mathcal{L}_{\text{smooth}} = -\log \frac{1}{L} \sum_{i=1}^{L} \frac{\sum_{j \in \mathcal{P}_i} f(\mathbf{a}_i, \mathbf{a}_{j})}{\sum_{l \in \mathcal{I}} f(\mathbf{a}_i, \mathbf{a}_{l})},
\end{equation}
Where $\mathcal{I}$ is the set of $L$ frame indices sampled at random, $\mathcal{P}_i$ is a subset of $\mathcal{I}$ indices that form a positive pair with $i$, \ie lie in the positive neighborhood of $M$ frames, and $f(x,z) = \exp(\cos(x, z) / \gamma)$, where $\gamma=0.03$ is a scaling temperature.

\begin{table}[t]
\centering
\resizebox{0.99\columnwidth}{!}{
	\begin{tabular}{l|llll|llll}
	\hline
	\multirow{2}{*}{Method} & \multicolumn{4}{c|}{Unsup. Segmentation} & \multicolumn{4}{c}{Zero-shot Localization}  \\
		& F1 & Prec. & Rec. & MoF  & IoU & Prec. & Rec. & MoF \\ 
		\hline
	Ours (24 slots)	& 27.9 & 21.4 & 40.5 &  \textbf{45.3} & 21.6 & 30.9 & 44.2 & 66.4 \\
	Ours (48 slots)	& 26.5 & 20.7 & 39.8 &  40.6 & \textbf{24.2} & \textbf{34.1} & 40.1 & 67.7 \\
	Ours (2 layers)	& 23.2 & 18.4 & 34.1 &  39.6 & 14.9 & 24.9 & 26.8 & \textbf{68.1} \\
	Ours (4 layers)	& 27.8 & 21.4 & 41.2 &  43.6 & 20.8 & 28.9 & 38.8 & 67.4 \\
	Ours (8 layers)	& 26.6 & 22.0 & 39.7 &  42.5 & 22.7 & 32.4 & \textbf{45.5} & 66.7 \\
		\hline
    Ours (6 layers, 32 slots) & \textbf{28.3} & \textbf{22.1} & \textbf{42.0} & 41.9 & 23.7 & 32.9 & 43.1 & 67.1 \\
		\hline
	\end{tabular}
	}
	\caption{Ablation study of StepFormer's training and inference components on CrossTask.} \label{tab:ablation}
	\vspace{-15pt}
\end{table}

\section{Baselines}\label{sec:baselines}
To verify the effectiveness of StepFormer, we compare our model to three baselines: Kukleva et al.~\cite{kukleva2019unsupervised}, Elhamifar et al.~\cite{elhamifar2020self}, and Shen et al.~\cite{Shen2021}.
We consider all the baselines as weakly-supervised, as they use information about the video task label during training.
Kukleva et al.~\cite{kukleva2019unsupervised} and Elhamifar et al.~\cite{elhamifar2020self} train a model purely from video;
thus, we use these baselines only in unsupervised step localization (Section 4.2 of the main paper).
More similar to StepFormer, Shen et al.~\cite{Shen2021} extract prototypes from a video on the fly and supervises them with text subtitles.
In principle, their prototypes should follow the temporal order and be alignable with text features, similar to our step slots.
Hence, we compare StepFormer to Shen et al. also in the zero-shot step localization setup (Section 4.3 of the main paper).
While Elhamifar et al. directly uses video task labels for supervision, Kukleva et al. and Shen et al. use such labels implicitly, \ie they train a separate model for each task using only the videos that belong to that task.
In this work, we adapt the methods of Kukleva et al.\ and Shen et al.\ to completely unsupervised training (\ie without video labels), by merging all tasks (and their videos) into a single dataset-level task, thereby not revealing the task-specific labels during training.
Unlike StepFormer, all the baseline methods must be trained and tested on the same dataset (as they learn task-specific step prototypes), and cannot generalize to new data.
For direct comparison with StepFormer, we train all the baselines using the same video features~\cite{Luo2020UniVL}, and fix the same training and testing splits in every dataset.

\section{Ablation study}\label{sec:ablation}
In this section, we complement Section 4.4 of the main paper with additional ablations of StepFormer components on the CrossTask dataset.
Specifically, we vary the number of transformer decoder layers and the number of output step slots used to describe a video, and report the results in Table~\ref{tab:ablation}.
First, as the table demonstrates, increasing the number of transformer layers to 6, \ie the default value used in the main paper, helps improve the performance. However, further increasing the number of layers actually hurts.
We attribute this effect to optimization difficulty of larger models.
Second, using 32 step slots to solve unsupervised step localization seems to be optimal for unsupervised step discovery and localization. However, to solve zero-shot step localization, more step slots seem to work better.
We attribute the increased zero-shot step localization performance with 48 slots to the improved text-to-slot matching step, as 48 slots offer more freedom in the matching.
Nevertheless, we elect to use the setup consisting of 6 layers and 32 step slots as it offers a good compromise in terms of performance on both target tasks.

{\small
\bibliographystyle{ieee_fullname}
\bibliography{bibref}

\begin{thebibliography}{10}\itemsep=-1pt

\bibitem{bi2021procedure}
Jing Bi, Jiebo Luo, and Chenliang Xu.
\newblock Procedure planning in instructional videos via contextual modeling
  and model-based policy learning.
\newblock In {\em Proceedings of the International Conference on Computer
  Vision (ICCV)}, 2021.

\bibitem{CaoJCCN20}
Kaidi Cao, Jingwei Ji, Zhangjie Cao, Chien{-}Yi Chang, and Juan~Carlos Niebles.
\newblock Few-shot video classification via temporal alignment.
\newblock In {\em Proceedings of the IEEE Conference on Computer Vision and
  Pattern Recognition (CVPR)}, 2020.

\bibitem{DETR}
Nicolas Carion, Francisco Massa, Gabriel Synnaeve, Nicolas Usunier, Alexander
  Kirillov, and Sergey Zagoruyko.
\newblock End-to-end object detection with transformers.
\newblock In {\em Proceedings of the European Conference on Computer Vision
  (ECCV)}, 2020.

\bibitem{ChangHS0N19}
Chien{-}Yi Chang, De{-}An Huang, Yanan Sui, Li Fei{-}Fei, and Juan~Carlos
  Niebles.
\newblock {D3TW}: {D}iscriminative differentiable dynamic time warping for
  weakly supervised action alignment and segmentation.
\newblock In {\em Proceedings of the IEEE Conference on Computer Vision and
  Pattern Recognition (CVPR)}, 2019.

\bibitem{procedure2020}
Chien-Yi Chang, De-An Huang, Danfei Xu, Ehsan Adeli, Li Fei-Fei, and
  Juan~Carlos Niebles.
\newblock Procedure planning in instructional videos.
\newblock In {\em Proceedings of the European Conference on Computer Vision
  (ECCV)}, 2020.

\bibitem{Chang2021}
Xiaobin Chang, Frederick Tung, and Greg Mori.
\newblock Learning discriminative prototypes with dynamic time warping.
\newblock In {\em Proceedings of the IEEE Conference on Computer Vision and
  Pattern Recognition (CVPR)}, 2021.

\bibitem{softdtw}
Marco Cuturi and Mathieu Blondel.
\newblock Soft-{DTW}: {A} differentiable loss function for time-series.
\newblock In {\em International Conference on Machine Learning (ICML)}, 2017.

\bibitem{DingX18}
Li Ding and Chenliang Xu.
\newblock Weakly-supervised action segmentation with iterative soft boundary
  assignment.
\newblock In {\em Proceedings of the IEEE Conference on Computer Vision and
  Pattern Recognition (CVPR)}, 2018.

\bibitem{dropdtw}
Nikita Dvornik, Isma Hadji, Konstantinos~G Derpanis, Animesh Garg, and Allan
  Jepson.
\newblock {Drop-DTW}: Aligning common signal between sequences while dropping
  outliers.
\newblock In {\em Advances in Neural Information Processing Systems (NeurIPS)},
  2021.

\bibitem{graphdropdtw}
Nikita Dvornik, Isma Hadji, Hai Pham, Dhaivat Bhatt, Brais Martinez, Afsaneh
  Fazly, and Allan Jepson.
\newblock Flow graph to video grounding for weakly-supervised multi-step
  localization.
\newblock In {\em Proceedings of the European Conference on Computer Vision
  (ECCV)}, 2022.

\bibitem{elhamifar2020self}
Ehsan Elhamifar and Dat Huynh.
\newblock Self-supervised multi-task procedure learning from instructional
  videos.
\newblock In {\em Proceedings of the European Conference on Computer Vision
  (ECCV)}, 2020.

\bibitem{caba2015activitynet}
Bernard~Ghanem Fabian Caba~Heilbron, Victor~Escorcia and Juan~Carlos Niebles.
\newblock {ActivityNet}: A large-scale video benchmark for human activity
  understanding.
\newblock In {\em Proceedings of the IEEE Conference on Computer Vision and
  Pattern Recognition (CVPR)}, 2015.

\bibitem{Girdhar2021}
Rohit Girdhar and Kristen Grauman.
\newblock {Anticipative Video Transformer}.
\newblock In {\em Proceedings of the International Conference on Computer
  Vision (ICCV)}, 2021.

\bibitem{d2tw}
Isma Hadji, Konstantinos~G. Derpanis, and Allan~D. Jepson.
\newblock Representation learning via global temporal alignment and
  cycle-consistency.
\newblock In {\em Proceedings of the IEEE Conference on Computer Vision and
  Pattern Recognition (CVPR)}, 2021.

\bibitem{TAN}
Tengda Han, Weidi~Xie Xie, and Andrew Zisserman.
\newblock Temporal alignment networks for long-term video.
\newblock In {\em Proceedings of the IEEE Conference on Computer Vision and
  Pattern Recognition (CVPR)}, 2022.

\bibitem{HuangFN16}
De{-}An Huang, Li Fei{-}Fei, and Juan~Carlos Niebles.
\newblock Connectionist temporal modeling for weakly supervised action
  labeling.
\newblock In {\em Proceedings of the European Conference on Computer Vision
  (ECCV)}, 2016.

\bibitem{joseph2021towards}
KJ Joseph, Salman Khan, Fahad~Shahbaz Khan, and Vineeth~N Balasubramanian.
\newblock Towards open world object detection.
\newblock In {\em Proceedings of the IEEE Conference on Computer Vision and
  Pattern Recognition (CVPR)}, 2021.

\bibitem{kukleva2019unsupervised}
Anna Kukleva, Hilde Kuehne, Fadime Sener, and Jurgen Gall.
\newblock Unsupervised learning of action classes with continuous temporal
  embedding.
\newblock In {\em Proceedings of the IEEE Conference on Computer Vision and
  Pattern Recognition (CVPR)}, 2019.

\bibitem{liang2022self}
Hanwen Liang, Niamul Quader, Zhixiang Chi, Lizhe Chen, Peng Dai, Juwei Lu, and
  Yang Wang.
\newblock Self-supervised spatiotemporal representation learning by exploiting
  video continuity.
\newblock In {\em AAAI Conference on Artificial Intelligence}, 2022.

\bibitem{locatello2020object}
Francesco Locatello, Dirk Weissenborn, Thomas Unterthiner, Aravindh Mahendran,
  Georg Heigold, Jakob Uszkoreit, Alexey Dosovitskiy, and Thomas Kipf.
\newblock Object-centric learning with slot attention.
\newblock {\em Advances in Neural Information Processing Systems (NeurIPS)},
  2020.

\bibitem{lu2022set}
Zijia Lu and Ehsan Elhamifar.
\newblock Set-supervised action learning in procedural task videos via pairwise
  order consistency.
\newblock In {\em Proceedings of the IEEE Conference on Computer Vision and
  Pattern Recognition (CVPR)}, 2022.

\bibitem{Luo2020UniVL}
Huaishao Luo, Lei Ji, Botian Shi, Haoyang Huang, Nan Duan, Tianrui Li, Jason
  Li, Taroon Bharti, and Ming Zhou.
\newblock {U}ni{VL}: A unified video and language pre-training model for
  multimodal understanding and generation.
\newblock {\em arXiv preprint arXiv:2002.06353}, 2020.

\bibitem{MaFK16}
Minghuang Ma, Haoqi Fan, and Kris~M. Kitani.
\newblock Going deeper into first-person activity recognition.
\newblock In {\em Proceedings of the IEEE Conference on Computer Vision and
  Pattern Recognition (CVPR)}, 2016.

\bibitem{miech2020end}
Antoine Miech, Jean-Baptiste Alayrac, Lucas Smaira, Ivan Laptev, Josef Sivic,
  and Andrew Zisserman.
\newblock {E}nd-to-{E}nd {L}earning of {V}isual {R}epresentations from
  {U}ncurated {I}nstructional {V}ideos.
\newblock In {\em Proceedings of the IEEE Conference on Computer Vision and
  Pattern Recognition (CVPR)}, 2020.

\bibitem{miech19howto100m}
Antoine Miech, Dimitri Zhukov, Jean-Baptiste Alayrac, Makarand Tapaswi, Ivan
  Laptev, and Josef Sivic.
\newblock How{T}o100{M}: {L}earning a {T}ext-{V}ideo {E}mbedding by {W}atching
  {H}undred {M}illion {N}arrated {V}ideo {C}lips.
\newblock In {\em Proceedings of the International Conference on Computer
  Vision (ICCV)}, 2019.

\bibitem{oord2018representation}
Aaron van~den Oord, Yazhe Li, and Oriol Vinyals.
\newblock Representation learning with contrastive predictive coding.
\newblock {\em arXiv preprint arXiv:1807.03748}, 2018.

\bibitem{radford2021learning}
Alec Radford, Jong~Wook Kim, Chris Hallacy, Aditya Ramesh, Gabriel Goh,
  Sandhini Agarwal, Girish Sastry, Amanda Askell, Pamela Mishkin, Jack Clark,
  et~al.
\newblock Learning transferable visual models from natural language
  supervision.
\newblock In {\em International Conference on Machine Learning}, pages
  8748--8763. PMLR, 2021.

\bibitem{richard2018neuralnetworkviterbi}
Alexander Richard, Hilde Kuehne, Ahsan Iqbal, and Juergen Gall.
\newblock {NeuralNetwork-Viterbi}: A framework for weakly supervised video
  learning.
\newblock In {\em Proceedings of the IEEE Conference on Computer Vision and
  Pattern Recognition (CVPR)}, 2018.

\bibitem{Sener2019}
Fadime Senner and Angela Yao.
\newblock Zero-shot anticipation for instructional activities.
\newblock In {\em Proceedings of the International Conference on Computer
  Vision (ICCV)}, 2019.

\bibitem{tcn}
Pierre Sermanet, Corey Lynch, Yevgen Chebotar, Jasmine Hsu, Eric Jang, Stefan
  Schaal, and Sergey Levine.
\newblock Time-contrastive networks: Self-supervised learning from video.
\newblock In {\em IEEE International Conference on Robotics and Automation
  (ICRA)}, 2018.

\bibitem{Shen2021}
Yuhan Shen, Lu Wang, and Ehsan Elhamifar.
\newblock Learning to segment actions from visual and language instructions via
  differentiable weak sequence alignment.
\newblock In {\em Proceedings of the IEEE Conference on Computer Vision and
  Pattern Recognition (CVPR)}, 2021.

\bibitem{sohn2016improved}
Kihyuk Sohn.
\newblock Improved deep metric learning with multi-class n-pair loss objective.
\newblock {\em Advances in Neural Information Processing Systems (NeurIPS)},
  29, 2016.

\bibitem{spacy}
spaCy.
\newblock Industrial-strength natural language processing.
\newblock \url{https://spacy.io/}.

\bibitem{srivastava2014dropout}
Nitish Srivastava, Geoffrey Hinton, Alex Krizhevsky, Ilya Sutskever, and Ruslan
  Salakhutdinov.
\newblock Dropout: a simple way to prevent neural networks from overfitting.
\newblock {\em The journal of machine learning research (JMLR)}, 2014.

\bibitem{COIN}
Yansong Tang, Dajun Ding, Yongming Rao, Yu Zheng, Danyang Zhang, Lili Zhao,
  Jiwen Lu, and Jie Zhou.
\newblock {COIN}: A large-scale dataset for comprehensive instructional video
  analysis.
\newblock In {\em Proceedings of the IEEE Conference on Computer Vision and
  Pattern Recognition (CVPR)}, 2019.

\bibitem{Tilk2016BidirectionalRN}
Ottokar Tilk and Tanel Alum{\"a}e.
\newblock Bidirectional recurrent neural network with attention mechanism for
  punctuation restoration.
\newblock In {\em INTERSPEECH}, 2016.

\bibitem{transformer}
Ashish Vaswani, Noam Shazeer, Niki Parmar, Jakob Uszkoreit, Llion Jones,
  Aidan~N Gomez, {\L}ukasz Kaiser, and Illia Polosukhin.
\newblock Attention is all you need.
\newblock {\em Advances in Neural Information Processing Systems (NeurIPS)},
  30, 2017.

\bibitem{xiong2020layer}
Ruibin Xiong, Yunchang Yang, Di He, Kai Zheng, Shuxin Zheng, Chen Xing,
  Huishuai Zhang, Yanyan Lan, Liwei Wang, and Tieyan Liu.
\newblock On layer normalization in the transformer architecture.
\newblock In {\em International Conference on Learning Representations (ICLR)},
  2020.

\bibitem{yang-hal-2021}
Antoine Yang, Antoine Miech, Josef Sivic, Ivan Laptev, and Cordelia Schmid.
\newblock {Just Ask: Learning to Answer Questions from Millions of Narrated
  Videos}.
\newblock In {\em Proceedings of the International Conference on Computer
  Vision (ICCV)}, 2021.

\bibitem{p3iv}
He Zhao, Isma Hadji, Nikita Dvornik, Konstantinos~G Derpanis, Richard~P Wildes,
  and Allan Jepson.
\newblock {P3IV}: Probabilistic procedure planning from instructional videos
  with weak supervision.
\newblock In {\em Proceedings of the IEEE Conference on Computer Vision and
  Pattern Recognition (CVPR)}, 2022.

\bibitem{YouCook2}
Luowei Zhou, Chenliang Xu, and Jason~J. Corso.
\newblock Towards automatic learning of procedures from web instructional
  videos.
\newblock In {\em AAAI Conference on Artificial Intelligence}, 2018.

\bibitem{CrossTask}
Dimitri Zhukov, Jean-Baptiste Alayrac, Ramazan~Gokberk Cinbis, David Fouhey,
  Ivan Laptev, and Josef Sivic.
\newblock Cross-task weakly supervised learning from instructional videos.
\newblock In {\em Proceedings of the IEEE Conference on Computer Vision and
  Pattern Recognition (CVPR)}, 2019.

\end{thebibliography}
}

\end{document}